\documentclass[12pt]{amsart}
\usepackage{ljm-auth}
\usepackage{url}

\newtheorem{definition}{Definition}

\tit{Mathematical Knowledge Representation:  Semantic Models and
Formalisms}
\shorttit{Mathematical Knowledge Representation} 

\author{Alexander M.~Elizarov,  Alexander V.~Kirillovich, Evgeny K.~Lipachev, Olga A.~Nevzorova, Valery D.~Solovyev, Nikita G.~Zhiltsov}
\crauthor{A.\,Elizarov et al.} 

\begin{document}

\maketit

\address{Kazan Federal University, N.~I.~Lobachevskii Institute of Mathematics and Mechanics,  Higher Institute of Information Technologies and Information Systems,
Kremlevskaya Str. 18, 420008, Kazan, Russia;\\
 Research Institute of Applied Semiotics of the Tatarstan Academy of Sciences, Baumana  Str. 20, 420111, Kazan, Russia
}

\email {\{amelizarov, alik.kirillovich, elipachev, onevzoro\}@gmail.com, maki.solovyev@mail.ru, nikita.zhiltsov@gmail.com}

\abstract{The paper provides a survey of semantic methods for
solution of fundamental tasks in mathematical knowledge management.
Ontological models and formalisms are discussed. We propose an
ontology of mathematical knowledge, covering a wide range of fields of
mathematics. We demonstrate applications of this representation in
mathematical formula search, and learning.}

\notes{0}{
\thank{This work was funded by the subsidy allocated to Kazan
Federal University for the state assignment in the sphere of
scientific activities (project 3056)}}

\section{Introduction}\label{intro}

The rapid growth of the modern science requires effective purpose-built information systems. Since inception of the first scientific information systems,
mathematicians have been involved in the full cycle of software product development, from idea to implementation.
A well-known example is \TeX{}, an open source typesetting system designed and mostly written by Donald Knuth~\cite{knuth}.
\TeX{} has a solid community of developers, researchers, and enthusiasts, who contribute new packages~\cite{ctan}.
The reader is likely aware of Mathematica~\cite{mathematica-site} and WolframAlpha~\cite{WolframAlpha-site} commercial systems,
led by a mathematician and physicist Stephen Wolfram according to his principles of computational knowledge theory (see e.g.~\cite{wolfram}).
Tools for mathematical content management are developed with the help of communities of mathematicians, e.g. MathJax~\cite{mathjax,cervone}
by American Mathematical Society, as well as independent researchers, e.g. ASCIIMathML \cite{ascii}. Math-Net.Ru~\cite{Zhizhchenko},
a collection of publications  from refereed journals, and arXiv.org, a collection of publicly available pre-prints,
are information systems that benefit from contributions of the mathematical community. The similar situation can be seen in other natural sciences.
For example, there are examples of information systems developed by chemists~\cite{cml,murray}.
However, the contemporary science community clearly lacks information systems, covering all its needs.

Main challenges in mathematical knowledge management (MKM) are
discussed in~\cite{mkm} --
\cite{nti:2014}.
Further, we frame the most
urgent tasks:
\begin{itemize}
  \item modeling representations of mathematical knowledge, i.e., techniques for representing MKM include data structures, logics, formal theories, diagrams;
  \item presentation formats, i.e.,
formats, programming languages etc.;
  \item authoring languages and tools;
  \item creating repositories of formalized mathematics, and mathematical digital libraries;

  \item mathematical search and retrieval, i.e., querying collections of mathematical documents;
  \item implementing math assistants, tutoring and assessment systems;
  \item developing collaboration tools for mathematics;
  \item creating new tools for detecting repurposing material, including plagiarism of others' work and self-plagiarism;
  \item creation of ``live documents''~\cite{parinov};
\item creation of interactive documents, e.g. efforts of the Liber Mathematicae community~\cite{liber-site,liber}
and Computable Document Format (CDF)~\cite{cdf} by Wolfram;
\item developing deduction systems, i.e.,
theorem provers and computer algebra systems (e.g. ~\cite{mizar,coq}). The solution of this task requires rigid formalization of mathematical statements and proofs.
\end{itemize}

While mathematics is full of formalisms, there is currently no a
single widely accepted formalism for computer mathematics. To
tackle this issue, we describe an approach that is based on Semantic
Web models and technologies~\cite{tim}. At the core of integration
of mathematical resources, there is building structured
representation of the scientific content. World Wide Web Consortium
(W3C) (www.w3.org) is an international community to develop standard
and technologies of Semantic Web, including special purpose markup
languages for many domains.

In this paper, we elaborate semantic-based approaches to solve some of the tasks described above.
In Section~\ref{sem}, we outline existing semantic models for mathematical documents.
In Section~\ref{formalizm}, we present $OntoMath^{PRO}$, a novel ontological model for mathematics that was developed by the authors
together with mathematicians from Kazan Federal University. Section~\ref{applicat} contains concrete applications  in search as well as
education powered by the ontology.

\section{Semantic Models of Mathematical Documents} \label{sem}

In this section, we give an overview of state-of-the-art semantic models of mathematical documents.

\subsection{Semantic markup for formulas}

Semantic markup enables automatic intelligent information processing. For representation of mathematical formulas,
there has been developed Mathematical Markup Language (MathML)~\cite{mathml}.
MathML was designed by W3C as a machine-readable language to both present and consume mathematical content in WWW.
The increasing role of this language in mathematical content management is discussed in~\cite{miner}.

Widely used tools for authoring mathematical articles include \LaTeX{}-based integrated development environments and office packages with mathematical formula support, such as MS Word+MathType. MathML Word2TeX~\cite{word2tex} and
\LaTeX{}ML~\cite{latex2} can be leveraged to convert documents from popular formats to XHTML+MathML for publication in Web.

\subsection{High-level models}

Open Mathematical Documents (OMDoc)~\cite{kohlhase:2006}, an XML-based language, is integrated with MathML/OpenMath and adds support of statements, theories,
and rhetorical structures to formalize mathematical documents. OMDoc has been used for interaction between structured specification systems and automated
theorem provers~\cite{omdoc:provers}. The OMDoc OWL Ontology (available at \url{http://kwarc.info/projects/docOnto/omdoc.html})
is based on the notion of statements. Sub-statement structures include definitions, theorems, lemmas, corollaries, proof steps.
The relation set comprises of partonomic (whole-part), logical dependency, and verbalizing properties.
The paper~\cite{ zhiltsov:2010} presents an OMDoc-based approach to author mathematical lecture notes using
S\TeX{} macro package~\cite{zhiltsov:2010, kohlhase:2005, kohlhase:2008}
in \LaTeX{} and expose them as Linked Data accessible in Web.
S\TeX{} offers macros for introducing new mathematical symbols and using arbitrary
metadata vocabularies. S\TeX{} is integrated with OMDoc ontology, providing definitions of OpenMath symbols and elements of the logical
structure of mathematical documents, such as theorems and proofs. This model also makes such documents directly available from the Web converting them
to XHTML/ RDFa format and offers different types of services
like notation explanation, versioning and semantic search.

The MathLang Document Rhetorical (DRa) Ontology~\cite{kamareddine} characterizes document structure elements according to their mathematical
rhetorical roles that are similar to the ones defined in the statement level of OMDoc. This semantics focuses on formalizing proof skeletons
for generation proof checker templates.

The Mocassin Ontology~\cite{solovyev} encompasses many structural elements of the models mentioned above. However, this model is more oriented on
representing structural elements and relations between them, e.g. logical dependency or referencing, occuring frequently in published scholarly papers in mathematics.
In~\cite{solovyev} we demonstrate its utility in the information extraction scenario.

\subsection{Terminological resources}

Terminological resources, such as vocabularies, datasets, thesauri, and ontologies include descriptions of mathematical knowledge objects.

The general-purpose DBpedia dataset~\cite{dbpedia} contains, according to our estimates, about 7,800 concepts (including 1,500 concepts with labels in Russian)
from algebra, 46,000 (9,200) concepts from geometry, 30,000 (4,300) concepts from mathematical logic, 150,000 (28,000) from mathematical analysis, and 165,000 (39,000)
concepts on theory of probability and statistics.

The ScienceWISE project~\cite{sciencewise:demo} gives over 2,500 mathematical definitions, including concepts from mathematical physics, connected with subclass-of,
whole-part, associative, and importance relationships.

The Online Encyclopedia of Integer Sequences~\cite{oeis} is a knowledge base of facts about numbers. Given a sequence of integers, this service (\url{http://oeis.org})
displays the information about its name, general formula, implementation in programming languages, successive numbers, references, and other relevant information.

Cambridge Mathematical Thesaurus~\cite{cmt} contains a taxonomy of about 4,500 entities in 9 languages from the undergraduate level mathematics,
connected with logical dependency and associative relationships.

\section{Ontologies as Formalisms for Mathematical Knowledge Representation} \label{formalizm}

We introduce ontology-based formalisms for knowledge representation as well as our novel ontological model for mathematics.

\subsection{Basic terms} \label{base-onto}

Both knowledge representation and knowledge interchange between
information agents, such as researchers and information systems,
rely on a conceptualization~\cite{genesereth}. Each communication
agent has its own vocabulary to refer to elements of the
conceptualization. Therefore, discrepancy between agent protocols
can occurr for two reasons: i) agents may have different
conceptualizations; ii) they may have incompatible models of
languages, i.e., meanings of terms. Effective communication requires
a single conceptualization as well as the sharable vocabulary.
Ontologies suffice this requirement.

Improving the classical definition by T.Gruber~\cite{gruber}, the authors of~\cite{studer} define an ontology as
``a formal, explicit specification of a shared conceptualization''.

An ontology defines basic concepts and relations between them of a given domain and includes:
\begin{itemize}
\item classes
\item properties
\item restrictions.
\end{itemize}

Hence, we accept the formal approach to ontology definition given by
N.~Guarino according to formal semantics~\cite{guarino}.

\begin{definition}\label{extensionalrelational}
An extensional relational
structure is a tuple $S = (D,R)$ where
\begin{itemize}
  \item $D$ is a set called the universe of discourse
  \item $R$ is a set of relations on $D$.
\end{itemize}
\end{definition}

Let $W$ the set of world states (also called worlds, or possible worlds) for an area of interest.

\begin{definition}\label{conceptual-relation}

A conceptual relation (or intensional relation) $\rho^n$ of arity $n$ on $<D,W>$
is a total function $\rho^n : W \rightarrow 2^{D^n}$
from the set $W$ into the set of all $n$-ary
(extensional) relations on $D$.
\end{definition}

From Definition~\ref{conceptual-relation}, we can provide a formal definition of conceptualization.

\begin{definition}\label{conceptualization}
A conceptualization (or intensional relational structure) is a triple $C = (D,W, \mathfrak{R})$ with
\begin{itemize}
    \item $D$ a universe of discourse;
    \item $W$ a set of world states;
    \item $\mathfrak{R}$ a set of conceptual relations on the domain space $<D,W>$.
\end{itemize}
\end{definition}

 Ontological commitment establishes the proper meanings of vocabulary elements.
Let $\mathbf{L}$ be a first-order logical language with vocabulary
$\mathbf{V}$ and $\mathbf{C} = (D,W, \mathfrak{R})$,  a conceptualization.

\begin{definition}\label{commitment}
An ontological commitment (or intensional first order structure) for $\mathbf{L}$ is a tuple $\mathbf{K} = (\mathbf{C}, \mathfrak{I})$, where $\mathfrak{I}$
(called intensional interpretation function) is a total function $\mathfrak{I} : V \rightarrow D \cup \mathfrak{R}$ that maps each vocabulary symbol
of $\mathbf{V}$ to either an element of $D$ or an intensional relation belonging to the
set $\mathfrak{R}$.
\end{definition}

Let $I: V \rightarrow D \cup R$ be any function that maps vocabulary
to the union of elements and relations of the universe of discourse
(called extensional interpretation function), and $S$ is from
Definition~\ref{extensionalrelational}. An intended model is a model
that conforms the chosen ontological commitment, or formally

\begin{definition}\label{intended-models}
 A model $M = (S,I)$ is called an
intended model of $\mathbf{L}$ according to $\mathbf{K}$ if
\begin{enumerate}
    \item for all constant symbols $c \in \mathbf{V}$ we have $I(c) = \mathfrak{I}(c);$
    \item there exists a world state $w \in W$ such that, for each predicate symbol $v \in \mathbf{V}$
there exists an intensional relation $\rho \in \mathfrak{R}$ such
that $\mathfrak{I}(v) = \rho$ and $I(v) =\rho (w).$
\end{enumerate}

\end{definition}

Finally, the ontology is defined as follows:

\begin{definition}[Ontology]\label{ontology-def}
An ontology $\mathbf{O}_\mathbf{K}$ for ontological commitment $\mathbf{K}$ is a logical theory
consisting of a set of formulas of $\mathbf{L}$, constructed so that the set of its models matches as close as possible the set of intended models of
$\mathbf{L}$ according to
$\mathbf{K}$.
\end{definition}

The ontology can be expressed in various formalisms. The most ubiquitous languages are $F$-logic~\cite{angele}, and, particularly,
description logics languages~\cite{baader}. In practice, Web Ontology Language (OWL)~\cite{owl:primer},
a knowledge representation language founded on a description logic SHIQ, is the most used in the Semantic Web community.

\subsection{OntoMath$^{PRO}$}

$OntoMath^{PRO}$~\cite{kesw} is the first attempt to build an
ontology of mathematical knowledge objects according to principles
described above.

Hence, we apply formalisms from the previous section to mathematics. We assume that, in our case, the universe of discourse is mathematical objects
from scientific refereed  publications. The conceptualization for mathematics is principles for classication of objects according to their characteristics.
The vocabulary represents the mathematical terminology. The ontological commitment is meanings of mathematical terms widely accepted in the contemporary
mathematical community. Then, the ontology captures the accepted conceptualization and the ontological commitment.

The current version of $OntoMath^{PRO}$ contains concepts from the pre-selected fields of mathematics, such as number theory, set theory, algebra, analysis,
geometry, mathematical logic, discrete mathematics, theory of computation, differential equations, numerical analysis, probability theory, and statistics.
The ontology defines six relations, such as taxonomic relation, logical dependency, associative relation between objects, belongingness of objects to fields
of mathematics, and associative relation between problems and tasks.

Each mathematical concept is represented as a class in the ontology. The class has definitions both in Russian and English, relations with other classes,
and links to verified Semantic Web resources~\cite{dbpedia,sciencewise:demo}.

The current version of ontology has 3,449 classes, 3,627  taxonomic and 1,139 non-taxonomic relations.
We distinguish two hierarchies of classes: a taxonomy of the fields of mathematics and a taxonomy of mathematical knowledge objects. In the taxonomy of fields,
most fundamental fields, such as geometry and analysis, have been elaborated thoroughly. For example, there have been defined specific sub-fields of geometry:
analytic geometry, differential geometry, fractal geometry and others. There are three types of top level concepts in the taxonomy of mathematical knowledge objects:
i) basic metamathematical concepts, e.g. Set, Operator, Map, Function, Predicate etc;
ii) root elements of the concepts related to the particular fields of mathematics, e.g.  Element of Logics;
iii) common scientific concepts: Problem, Method, Statement,  and Formula. Concrete theoretical results, e.g. Arslanov's completeness criterion, can be found in lower levels.

\section{Applications}\label{applicat}

We present applications of the proposed semantic models for mathematical formula search and learning.

\subsection{Mathematical formula search}\label{search-math}

We have implemented two applications for mathematical formula search: syntactical search of formulas in MathML, and semantic ontology-based search.

The syntactical search leverages formula parts from documents formatted in \TeX{}. Our algorithm~\cite{elizarov:mathml} transforms formulas in \TeX{} to MathML.
We set up an information retrieval system prototype for a collection of articles in Lobachevskii Journal of Mathematics (LJM, \url{http://ljm.ksu.ru}).
For the end-user, the query input interface supports a convenient \TeX{} syntax. The search hit description includes hightlighted occurrences of formulas
as well as document metadata.

In our previous work~\cite{Nikita}, we have developed a semantic publishing platform for scientific collections in mathematics that analyzes the underlying
semantics in mathematical scholarly papers  and effectively builds their consolidated ontology-based representation. The current data set contains a semantic
representation of articles of ``Proceedings of Higher Education Institutions: Mathematics  journal''.

Our demo application (\url{http://cll.niimm.ksu.ru/mathsearch}) features a use case of querying mathematical formulas in the published dataset that are relevant
to a given mathematical concept. The supported user input is close to a keyword search: our system is agnostic to a particular symbolic notation used to express
mathematical concepts, and the user is able to select query suggestions by keywords. Our search interface also supports filtering by the document structure context,
i.e., a particular segment of the document (e.g. a theorem or a definition) that contains the relevant formula.

\subsection{Learning}\label{learning}
For a practicing mathematician, an ability of solving problems is crucial.
The proficient solver must realize relationships between particular methods, tasks, and proof techniques to make the transition from solving problems
to proving theorems~\cite{velleman}.
We describe our experiments on ontology-based assessment of the competence of students, who attended a course on numerical analysis.

For our experiments, we extracted a small fragment of $OntoMath^{PRO}$ ontology.
It contains taxonomies of tasks and solving methods for systems of linear equations (numerical analysis) as well as relationships between them.

The experiment participants were students who attended the course
and had high overall grades. Each participant is given a list of
classes and asked to link them using only two relationships:
taxonomic relation and \emph{solves}. Therefore, we treat this task as a
classification task. We use standard performance measures for
classification tasks, such as precision (P), recall (R), and F-score
$=2\cdot\frac{P*R}{P+R}$.

According to our results, reconstruction of concept properties is
the hardest task (35\% F-score on average) for most students
comparing to reconstruction of taxonomies (83\%). It means that the
ontology could be used by students to conceive the correct
conceptualization of a field of mathematics. The detailed analysis
of the experiments is provided in~\cite{kesw}.

\section{Conclusion}\label{conclusion}

The paper summarizes the key tasks in mathematical knowledge representation. We give an overview of state-of-the-art semantic models of mathematical documents.
We introduce ontology-based formalisms for knowledge representation as well as our novel ontological model, $OntoMath^{PRO}$, for mathematics.
We present applications of the proposed semantic models for mathematical formula search and learning.

 We emphasize that while the ontology has achieved maturity, it is the result of ongoing work. The ontology is publicly available on \url{ontomathpro.org}.
 On this webpage, we encourage our colleagues to take part in collaborative editing, including correction and contributing new classes, relations, and definitions.
 We also organize a discussion to prospect novel applications.

\textbf{Acknowledgments:} A.~Kirillovich would like to thank Evelina Khakimova (University of Virginia),
Claudia Acevedo (Lemoine Editores), and Maria Isabel Duarte (EAFIT University) for the assistance in the work with bibliographic sources.

\end{document}